\documentclass[conference]{IEEEtran}
\IEEEoverridecommandlockouts
\usepackage{algorithm,algorithmic}
\usepackage{cite}
\usepackage{fancyhdr}
\usepackage{amsmath,amssymb,amsfonts}
\usepackage{algorithmic}
\usepackage{graphicx}
\usepackage{textcomp}
\usepackage{xcolor}
\usepackage{hyperref}
\usepackage{float}
\usepackage{fancyhdr}
\usepackage{lastpage}
\usepackage{eso-pic}
\usepackage{fancyhdr}
\def\BibTeX{{\rm B\kern-.05em{\sc i\kern-.025em b}\kern-.08em
    T\kern-.1667em\lower.7ex\hbox{E}\kern-.125emX}}
\begin{document}

\title{Distinguishing Human Generated Text From ChatGPT Generated Text Using Machine Learning}

\author{Niful Islam$^1$, Debopom Sutradhar$^1$, Humaira Noor$^1$, \\ Jarin Tasnim Raya$^2$, Monowara Tabassum Maisha$^2$, Dewan Md Farid$^1$ \\
$^1$Department of CSE, United International University (UIU), Bangladesh\\
$^2$Department of CSE, University of Asia Pacific (UAP), Bangladesh\\
Email : \{nislam201057, dsutradhar201046\}@bscse.uiu.ac.bd, hnoor222007@mscse.uiu.ac.bd,  \\ \{20101002, 20101001\}@uap-bd.edu, dewanfarid@cse.uiu.ac.bd
}
\maketitle

\begin{abstract}
 ChatGPT is a conversational artificial intelligence that is a member of the generative pre-trained transformer of the large language model family. This text generative model was fine-tuned by both supervised learning and reinforcement learning so that it can produce text documents that seem to be written by natural intelligence. Although there are numerous advantages of this generative model, it comes with some reasonable concerns as well. This paper presents a machine learning-based solution that can identify the ChatGPT delivered text from the human written text along with the comparative analysis of a total of 11 machine learning and deep learning algorithms in the classification process. We have tested the proposed model on a Kaggle dataset consisting of 10,000 texts out of which 5,204 texts were written by humans and collected from news and social media. On the corpus generated by GPT-3.5, the proposed algorithm presents an accuracy of 77\%.
\end{abstract}


\section{Introduction}
The emergence of generative AI models are rapidly changing our way of communication. It is being extensively used in content creation, arts and design, healthcare and many more. Although these models, especially conversational AI models like ChatGPT, have the potential of revolutionizing the society, they also come with some possible dangers. One of the biggest concerns is that, it can produce false news or spread misinformation \cite{hacker2023regulating} \cite{botha2020fake}. Since, the AI-generated texts are almost identical to the human-generated texts, the model can be used to manipulate individuals or organizations in various ways. There are some legal and ethical concerns of using generative AI. Since these models are trained on large datasets, there might remain some bias in a particular sector. For decision making, if an individual employs these models, it may lead to discriminatory attitudes. Furthermore, students may rely too heavily on the AI generated tools, which could damage their critical thinking and communication ability, negatively impacting their academic and professional life \cite{qadir2022engineering}. For newly developed problems, it requires new solutions. As these models are trained on old data, asking solutions for advanced challenges might provide misleading solutions \cite{cao2023comprehensive}. In addition, incorporating conversational AI into the system could result in low user satisfaction. Therefore, it requires a system for identifying human generated text and AI generated text. \par

Natural Language Processing (NLP) is a rapidly growing field of study that works on understanding human language. NLP gives machines the ability to learn human language by turning it into numerical data \cite{khurana2023natural}. With the increasing number of digital texts, the need of NLP is growing rapidly\cite{feder2022causal}. In recent years, NlP has provided a large scale analysis and management of text data making sentiment analysis, emotion detection and other complicated tasks possible. Furthermore, with the help of NLP, it is possible to detect mental illness at early stage and provide treatment \cite{zhang2022natural}. Previously the training of NLP models were slow and inefficient \cite{chen2021crossvit}. Especially after 2017, the innovation of transformer architecture has revolutionized the NLP field. The transformers made NLP tasks to be carried out sequentially that gave birth to large language models like ChatGPT. This models are so efficient that it can imitate human behaviour which creates some reasonable concerns.\par 
Transformers are one of the most powerful tools for natural language processing \cite{gillioz2020overview}. It can largely be divided into two parts. The encoder and the decoder. The encoder part takes a text sequence as input and produces a sequence of encoded representation. The difference between other architecture is that the encoded sequence is more context aware. When a series of encoders are stacked togather, it is called a BERT \cite{singh2021nlp}. The decoder part, on the other hand, is able to generate an arbitrary length sequence. Stacking decoder blocks produces an architecture named GPT. ChatGPT is also a transformer based architecture that has been trained on a large set of public data in self-supervised fashion. It has more than a billion parameters making it on of the biggest language model available. It is considered a major breakthrough since it's release in November, 2022. \par

In this paper, we present a machine learning based approach for detecting ChatGPT generated text and human generated text. The proposed model involves vectoring sentences using TF-IDF vectorizer and then classifying it employing extremely randomized trees classifier. This article also presents a comparative analysis of different machine learning algorithms namely Logistic Regression, Support Vector Machine, Decision Tree, K- Nearest Neighbor, Random Forest, AdaBoost, Bagging Classifier and deep learning algorithms named Multi-layer Perceptron and Long Short-Term Memory  for detecting ChatGPT generated text along with the impact of some data pre-processing techniques in the classification process. To summarize, the article presents the followings: 
\begin{itemize}
    \item A machine learning based model for differentiating ChatGPT generated text from human generated text.
    \item Comparative analysis of different machine learning and deep learning algorithms in the classification process.
\end{itemize}
The article is organized such that Section \ref{sec:rw} contains the prior works for solving the same problems followed by our proposed method in Section \ref{sec:method}. Section \ref{sec:result} holds the outcomes obtained from the study. The article terminates in Section \ref{sec:conclusion}.

\section{Related Work}
\label{sec:rw}

To detect AI generated texts multiple approaches are proposed. Traditionally Sebastian Gehrmann et al. \cite{gehrmann2019gltr} proposed a statistical method to distinguish machine and human generated text. The paper introduces a tool named GLTR. The GLTR tool is a 6-gram character-level statistical language model model, which is trained on a large corpus of text data. The tool uses this model to calculate the probability of each character in the generated text, and then shows any character that has a low probability of occurring in the training corpus. Anton Bakhtin et al. \cite{bakhtin2019real} proposed a Energy-based model (EBM) to discriminate machine generated text. EBM is also a statistical model which finds an energy function from given data. The authors used a comparatively larger dataset collected from human to machine conversation. Eric Mitchell et al.\cite{mitchell2023detectgpt} proposed a zero shot learning method called DetectGPT. This method detects whether text is machine-written or not by calculating log probabilities computed by the model of interest. On text samples generated by the GPT-2, the research team conducted an  study \cite{solaiman2019release}. Atsumu Harada  et al. \cite{harada2021discrimination} gathered two datasets, one with sentences produced by humans, the other with sentences written by both humans and machines. The cosine similarity between sentence pairs was then calculated as a measure of text consistency. Finally, based on the cosine similarity ratings of the sentences, they classified them as either human-written or human and machine-written using machine learning methods. Sandra Mitrovic et al. \cite{mitrovic2023chatgpt} proposed a transformer based model to detect chatgpt generated texts. To determine if a text was produced by ChatGPT or a person, the paper's authors developed a machine learning methodology. The model was trained on a dataset of 10,000 text samples that were classified as either human- or ChatGPT-generated. It is based on a combination of text-based and user-based attributes. Tiziano Fagni et al. \cite{fagni2021tweepfake} proposed a method to detect deepFake tweets. At first tweets were generated by different language models. Then different machine learning methods were used with tf-idf and Bag of Words techniques.Sankar Sadasivan et al. \cite{sadasivan2023can}  evaluated the effectiveness of several existing approaches for detecting AI-generated text including rule-based methods, statistical methods, and machine learning-based methods.They discover that although these techniques can be effective in identifying some sorts of AI-generated text, they are frequently open to adversarial attempts that trick them into thinking the material is human-generated. A lightweight neural network-based paraphraser was developed and applied it to the AI-generated texts.John Kirchenbauer et al. \cite{kirchenbauer2023watermark} introduced a watermarking method. This method add a small amount of noise to the weights of the LLM during training.The noise is made with the intention of encoding a distinct watermark signal that can later be decoded by a watermark detector.The GPT-2 and GPT-3 language models are used to demonstrate the utility of their watermarking technology. The watermark can be found even after fine-tuning the LLM on fresh data and that it is resistant to a variety of attacks, including gradient masking and weight perturbations. In a paper Kalpesh Krishna et al.\cite{krishna2023paraphrasing}  created a substantial amount of AI-generated text samples using a number of cutting-edge language models, such as GPT-3 and T5. The efficiency of several rule-based and machine learning-based text-derived AI detectors is then assessed using the created samples. A retrieval-based defensive method was proposed that depends on determining the text's original author. The suggested technique operates by maintaining a database of known AI-generated text samples and their associated original sources, and by comparing any new text samples against this database to identify probable sources. The authors demonstrate that the suggested retrieval-based defensive mechanism is successful in identifying material that has been paraphrased by AI, with detection a good accuracy. Souradip Chakraborty et al. \cite{chakraborty2023possibilities} proposed multiple possibilities that can detect AI generated texts including some statistical methods and several machine learning algorithms. In most of the papers GPT-2 or previous version of GPT-3 were used. But in our paper we approached with conventional machine learning algorithms but with a dataset that was generated by GPT-3.5 which were more human like.

\section{Methodology}
\label{sec:method}
\begin{figure}
    \centering
    \includegraphics[width=\linewidth]{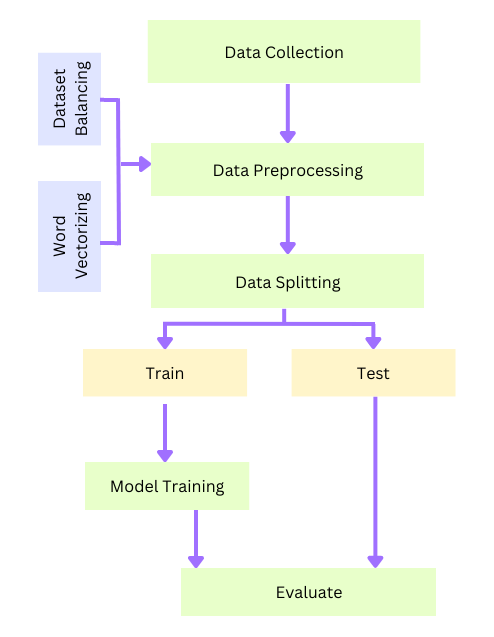}
    \caption{High level overview of the process}
    \label{fig:process}
\end{figure}

The aim of this research is to differentiate human text from generative model text using machine learning. In Figure \ref{fig:process}, the high level overview of our process is described. The task initiates by data collection. Section \ref{sec:data-collect} holds the detailed process of this stage. \par

In the data preprocessing stage, the dataset was balanced using undersampling technique. Moreover, the class column was converted into numerical values using binary encoding also known as one-hot encoding. Deleting stop words in the pre-processing stage impacted the classification performance negatively since the selection of stop words play a crucial role in differentiating human and AI. Finally, since machines can understand numbers only, the sentences were vectorized using TF-IDF vectorizer. The details of this technique is mentioned in section \ref{sec:tfidf}. The speciality of this vectorizer over other methods is its ability to capture the importance of a word. \par
\begin{figure}[H]
    \centering
    \includegraphics[width=\linewidth]{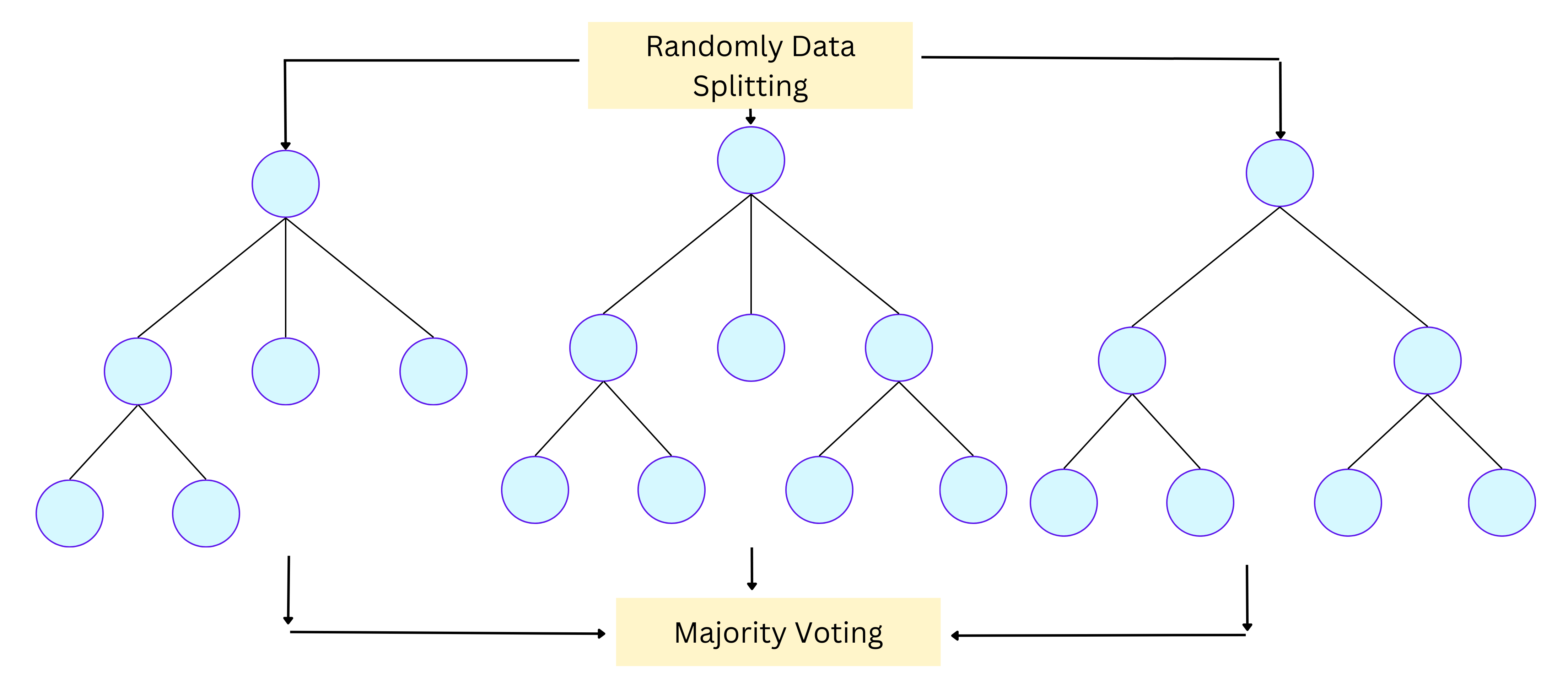}
    \caption{Extra Tree Classifier}
    \label{fig:etc}
\end{figure}

\begin{figure}[H]
    \centering
    \includegraphics[width=\linewidth]{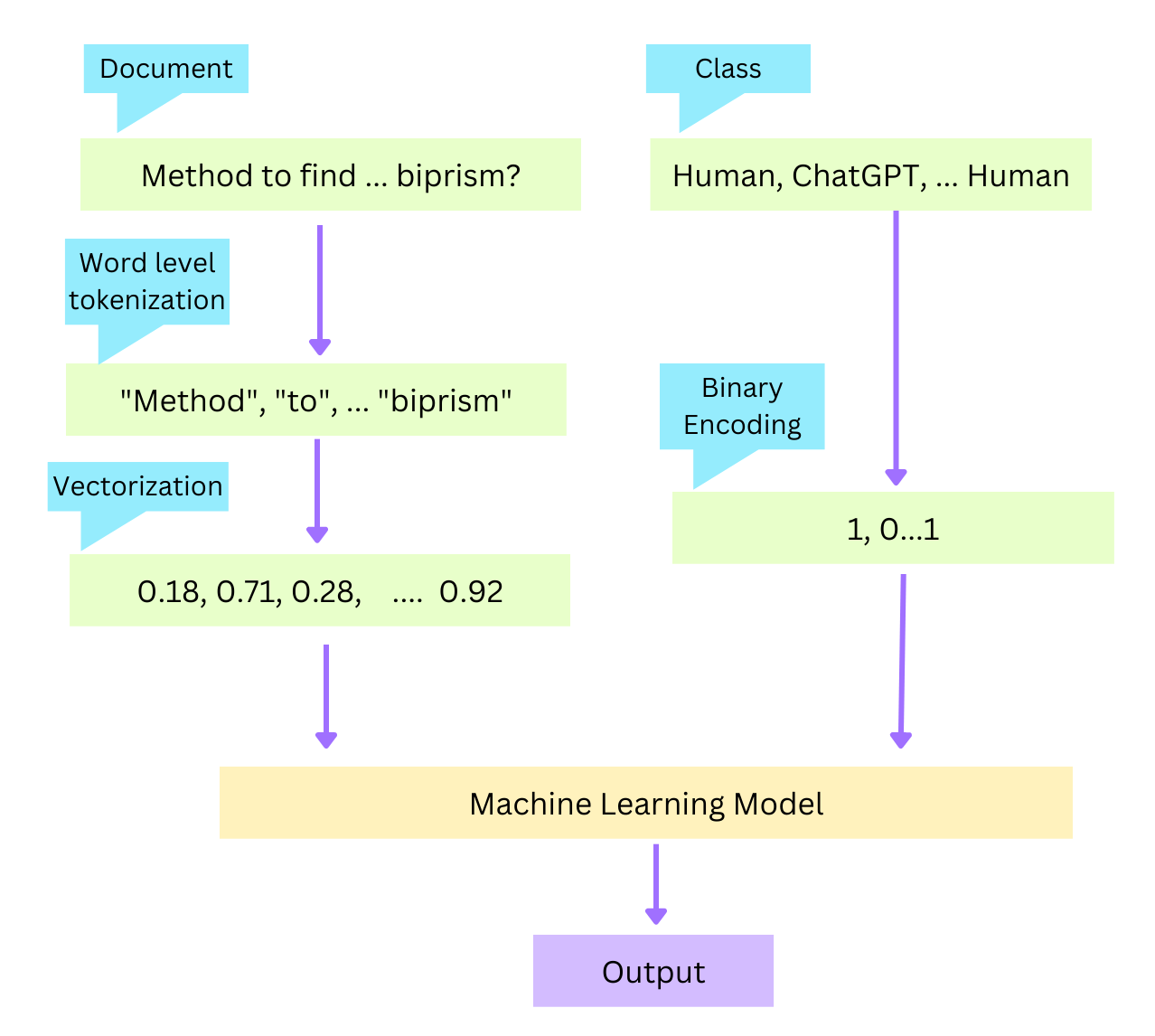}
    \caption{Detailed overview of the process}
    \label{fig:detailed}
\end{figure}

Later, the data was split into two halfs with a ratio of 80:20 where majority portion was kept for training and the rest for testing. Figure \ref{fig:detailed} presents the detailed description of the process. Total eleven models were selected for ablation study. Section \ref{sec:alg} presents a small description of every algorithms. The final model selected for this task was Extremely Randomized Trees Classifier (ERTC). ERTC is an ensemble algorithm that is based on decision tree. However, instead of selecting the best partition point, it splits the data based on random points. On the training phase, it constructs some number of decision trees based on randomly selected attributes and features \cite{ntahobari2022enhanced}. At testing, it takes majority voting for prediction. An illustration of ERTC is shown in Figure \ref{fig:etc}. There are several hyper-parametres of this algorithm. In our research, we have found that 50 Decision Tree classifiers without pruning works the best for this problem with splitting criteria as gini. 

\subsection{Data Collection and Preprocessing}
\label{sec:data-collect}
\begin{figure}[H]
    \centering
    \includegraphics[width=\linewidth]{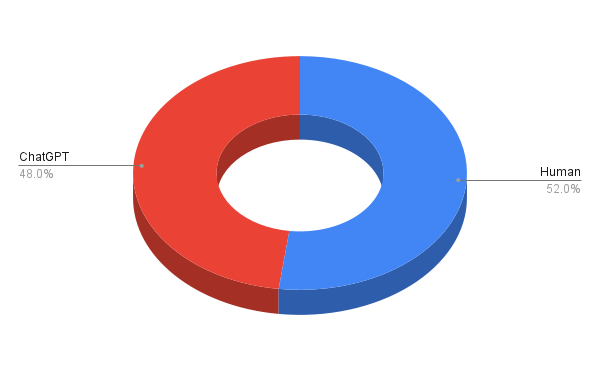}
    \caption{Dataset distribution }
    \label{fig:data-dist}
\end{figure}

The dataset consists of 10,000 texts among which 5204 texts are generated from humans and rest of them are from ChatGPT. Figure \ref{fig:data-dist} holds the distribution of the dataset. The initial dataset was constructed by first collecting data from Quora and CNN news using web scrapping. Later they were given to ChatGPT for paraphrasing. However, the initial datset had a 1:8 ratio of human:ChatGPT text. Later, data-preprocessing was applied to make the dataset balanced.

\subsection{TF-IDF Vectorizer}
\label{sec:tfidf}
Term Frequency-Inverse Document Frequency (TF-IDF ) is a popular word vectorization technique. The idea behind TF-IDF is to emphasize the important words. The process of calculating TF-IDF has two steps. First step is to calculate the term frequency, that is how often a term (word or group of words) appears in the document. Let term be denoted with $t$, document with $d$, the collection of documents (corpus) with $D$ and total number of documents in the corpus with $N$. Now, the term frequency is calculated by dividing the number of occurrences of a term in the document by the total number of terms. Equation \ref{eq:tf} holds the mathematical formula of calculating TF.
\begin{equation}
    \label{eq:tf}
    TF(t,d) = \frac{n_{t,d}}{\sum_{k} n_{k,d}}
\end{equation}

Inverse Document Frequency, on the other hand, measures how important a term is by calculating the frequency of that term in the corpus. As described in Equation \ref{eq:idf}, it is calculated by taking the logarithm of the total number of documents in the corpus divided by the number of documents that contain the term.

\begin{equation}
    \label{eq:idf}
    IDF(t,D) = \log \frac{N}{df(t)}
\end{equation}

Finally, the IF-IDF is calculated by multiplying the TF and IDF calculated in the steps above. 
\begin{equation}
    \label{eq:tfidf}
\text{TF-IDF}(t, d, D) = \text{TF}(t, d) \cdot \text{IDF}(t, D)
\end{equation}

\subsection{Algorithms}
\label{sec:alg}
\textbf{Logistic Regression:} Logistic regression is a statistical method which is mainly used for binary classification. It calculates relationship between input data and class variable to make prediction. \\
\textbf{Support Vector Machines:} Support Vector Machine (SVM) is used for both classification and regression. Finding a hyperplane in an N-dimensional space that clearly differentiates the data points is the goal of the SVM. \\
\textbf{Decision Tree:}  Decision trees constructs a tree data structure based on information obtained from the data. While inference, it traverses the tree to find the appropriate class.\\
\textbf{K- Nearest Neighbor:} The K-nearest neighbors algorithm (KNN) is based on the theory of proximity. It finds the class label of a given data by calculating the majority of the K nearest instances. \\
\textbf{Random Forest:} Random Forest is a supervised machine learning algorithm which combines some decision trees using the concept of attribute bagging. \\ 
\textbf{AdaBoost:} AdaBoost algorithm, also known as Adaptive Boosting, is a Boosting method used in machine learning as an ensemble method. It is more useful for noisy data.\\ 
\textbf{Bagging Classifier:} An ensemble meta-estimator called a bagging classifier fits base classifiers one at a time to random subsets of the original dataset, and it then averages or votes on each classifier's individual predictions to produce a final prediction.\\
\textbf{Gradient Boosting: } In order to minimize a loss function, the functional gradient algorithm known as Gradient Boosting repeatedly chooses a function that points in the direction of a weak hypothesis or a negative gradient. A powerful predicting model is created by the gradient boosting classifier by combining several weak learning models.\\
\textbf{Multi-layer Perceptron} : Artificial neurons are the main concept of Multi-layer Perceptron (MLP). These neurons are a set of interconnected units or nodes that loosely resemble the neurons in a biological brain. Like the synapses in a biological brain, each connection has the ability to send a signal to neighboring neurons. An artificial neuron can signal neurons that are connected to it after processing signals that are sent to it. The output of each neuron is calculated by some non-linear function of the sum of its inputs.\\
\textbf{Long Short-Term Memory: } Long Short-Term Memory(LSTM) is a type of recurrent neural network that can remember information of long sequences. \\
\textbf{Extremely Randomized Trees} : Extremely Randomized Trees, sometimes referred to as Extra Trees, build numerous trees, similar to Random Forest (RF) techniques, over the whole dataset during training. The difference between RF classifier is that RF splits data based on best splitting criteria but Extra Trees classifier splits data randomly. \\

\section{Results}
\label{sec:result}

\begin{table*}[!ht]
    \centering
    \caption{Performance of differnet classifiers}
    \begin{tabular}{|l|l|l|l|l|l|}
    \hline
        \textbf{Model} & \textbf{Accuracy} & \textbf{Precision} & \textbf{Recall} & \textbf{F1-Score} & \textbf{MCC}  \\ \hline
        Logistic Regression & 0.74 & 0.73 & 0.73 & 0.73 & 0.48  \\ \hline
        Support Vector Machines & 0.75 & 0.75 & 0.71 & 0.73 & 0.50  \\ \hline
        Decision Tree & 0.63 & 0.75 & 0.79 & 0.67 & 0.29  \\ \hline
        K- Nearest Neighbor & 0.69 & 0.67 & 0.68 & 0.67 & 0.37  \\ \hline
        Random Forest & 0.76 & 0.73 & 0.81 & 0.76 & 0.53  \\ \hline
        AdaBoost & 0.71 & 0.68 & 0.74 & 0.71 & 0.43  \\ \hline
        Bagging Classifier & 0.74 & 0.71 & 0.75 & 0.73 & 0.47  \\ \hline
        Gradient Boosting & 0.71 & 0.66 & 0.78 & 0.72 & 0.42  \\ \hline
        Multi-layer Perceptron & 0.72 & 0.73 & 0.72 & 0.72 & 0.43  \\ \hline
         Long Short-Term Memory & 0.73 & 0.73 & 0.77 & 0.75 & 0.46 \\ \hline
         \textbf{Extremely Randomized Trees} & \textbf{0.77} & \textbf{0.74} & \textbf{0.78} & \textbf{0.76} & \textbf{0.54} \\ \hline

    \end{tabular}
    \label{tab:performance}
\end{table*}

\subsection{Experimental Setup}
The experiment was carried out on a jupyter notebook and the machine was equipped with a CPU of ryzen 5 5600G. The CPU has an integrated Graphics Processing Unit (GPU) for carrying out deep learning tasks. Moreover, the machine also had 16 GB RAMs. Python was used as programming language along with four libraries named Numpy, Pandas, SKlearn and Tensorflow.\\

\subsection{Evaluation Matrices}
Evaluation matrices are used to measure the performance of the model. Different evaluation matrix provides different perspective of the result. In this paper, we have used five matrices namely accuracy, precision, recall,F1 score and Matthews correlation coefficient (MCC). Accuracy calculates the percentage of correctly predicted instances. As shown in Equation \ref{eq:acc}, it is calculated by the total number of correctly predicted samples divided by the total number of samples.\\ 

\begin{equation}
\label{eq:acc}
\mathrm{Accuracy} = \frac{\mathrm{TP} + \mathrm{TN}}{\mathrm{TP} + \mathrm{TN} + \mathrm{FP} + \mathrm{FN}}    
\end{equation}
While precision calculates the true positive predictions over all positive predictions, recall measures the true positive predictions out of all actual positive instances. The formula of precision and recall are demonstrated in Equation \ref{eq:prec} and Equation \ref{eq:recall} respectively. 
\begin{equation}
\label{eq:prec}
\mathrm{Precision} = \frac{\mathrm{TP}}{\mathrm{TP} + \mathrm{FP}}
\end{equation}

\begin{equation}
\label{eq:recall}
\mathrm{Recall} = \frac{\mathrm{TP}}{\mathrm{TP} + \mathrm{FN}}
\end{equation}

As shown in Equation \ref{eq:f1}, F1 score is the mean of precision and recall. This provides a more balanced measurement than precision and recall.
\begin{equation}
\label{eq:f1}
\mathrm{F1\text{-}score} = 2 \cdot \frac{\mathrm{Precision} \cdot \mathrm{Recall}}{\mathrm{Precision} + \mathrm{Recall}}
\end{equation}
Finally, the Matthews Correlation Coefficient (MCC) is a more stable matrix that takes all four coefficients into account. This matrix is more significant than all other matrices mentioned above \cite{chicco2020advantages}. The formula for calculating MCC is mentioned in Equation \ref{eq:mcc}.

\begin{equation}
\label{eq:mcc}
\mathrm{MCC} = \frac{\mathrm{TP} \cdot \mathrm{TN} - \mathrm{FP} \cdot \mathrm{FN}}{\sqrt{(\mathrm{TP} + \mathrm{FP})(\mathrm{TP} + \mathrm{FN})(\mathrm{TN} + \mathrm{FP})(\mathrm{TN} + \mathrm{FN})}}
\end{equation}

\begin{figure}[H]
    \centering
    \includegraphics[width=\linewidth]{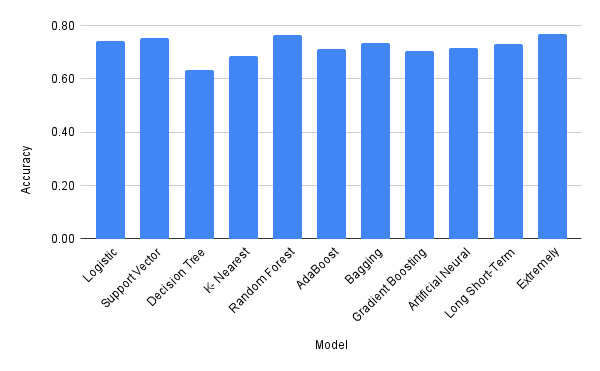}
    \caption{Accuracy of different classifiers}
    \label{fig:acc}
\end{figure}

\subsection{Experimental Results}
We have tested the performance against nine machine learning classifiers along with an MLP and LSTM model with different hyper-parameters. Table \ref{tab:performance} holds the detailed performance analysis of different models. Figure \ref{fig:acc} shows a diagrammatic accuracy comparison of different classifiers.

\begin{figure}[H]
    \centering
    \includegraphics[width=\linewidth]{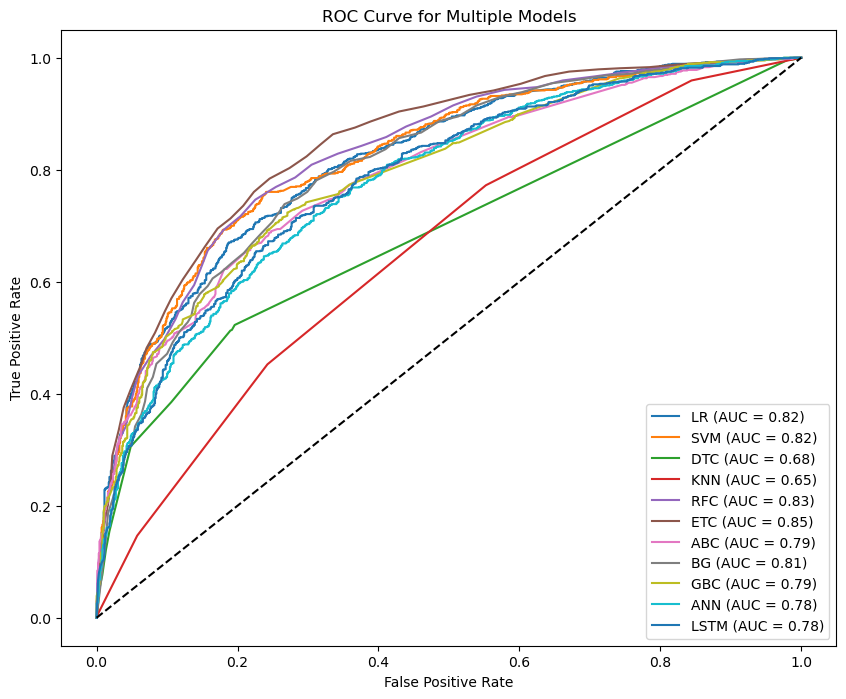}
    \caption{ROC curve of all models compared}
    \label{fig:roc}
\end{figure}
Form the results, it is clearly visible that extra tree classifier outperforms all the classifiers with an accuracy of 77\%. Moreover, it also has the highest MCC score that is the most informative evaluation matrix compared. The results also demonstrate that some well known classifier such as K-Nearest Neighbor and Decision Tree classifier performs poorly on this dataset. For deep learning based Artificial Neural Network and Long Short-Term Memory, it was trained on 15 epoch. Although it had a high training accuracy, it performed poorly on testing. Some regularization techniques may improve the performance. For further investigation of the result, Figure and \ref{fig:roc} presents roc curve. These results show that the model is not biased to a particular class.


\section{Conclusion}
\label{sec:conclusion}
In this research, we proposed a model that can differentiate between the text generated from human and ChatGPT. Since generating AI has become very advanced, it is difficult to distinguish between human text and machine generated text. However, in this paper, we have presented a machine learning based approach that can effectively identify two types of text. With the continuing research in this area  we expect to see more sophisticated models for solving this problem that will ensure transparency and accountability in the day to day life.

\bibliographystyle{IEEEtran}
\bibliography{FILES/bibfile.bib}

\end{document}